\documentclass[conference]{IEEEtran}
\IEEEoverridecommandlockouts

\usepackage{cite}
\usepackage{amsmath,amssymb,amsfonts}
\usepackage{algorithmic}
\usepackage{graphicx}
\usepackage{textcomp}
\usepackage{xcolor}
\usepackage{url}

\def\BibTeX{{\rm B\kern-.05em{\sc i\kern-.025em b}\kern-.08em
    T\kern-.1667em\lower.7ex\hbox{E}\kern-.125emX}}

\usepackage[compatibility=false]{caption}
\usepackage{subcaption}
\usepackage{enumitem}
\usepackage{booktabs}
\graphicspath{{.}}

\renewcommand{\it}[1]{\textit{#1}}
\newcommand{\bt}[1]{\textbf{#1}}
\newcommand{\hide}[1]{\ignorespaces}

\begin{document}

\title{Fusing Cellular Network Data and Tollbooth Counts for Urban Traffic Flow Estimation
    \\ \thanks{This research was funded by the PERSEUS project under the Marie Skłodowska-Curie grant agreement No. 101034240.}
}
\author{
    \IEEEauthorblockN{Oluwaleke Yusuf}
    \IEEEauthorblockA{\textit{Department of Engineering Cybernetics} \\
        \textit{Norwegian University of Science and Technology (NTNU)} \\
        NO-7491 Trondheim, Norway \\
        ORCID: 0000-0002-5904-648X}
    \and
    \IEEEauthorblockN{Shaira Tabassum}
    \IEEEauthorblockA{\textit{Department of Computer Science} \\
        \textit{Norwegian University of Science and Technology (NTNU)} \\
        NO-7491 Trondheim, Norway \\
        ORCID: 0000-0001-8374-5195}
}
\maketitle

\begin{abstract}
    Traffic simulations, essential for planning urban transit infrastructure interventions, require vehicle-category-specific origin--destination (OD) data. Existing data sources are imperfect: sparse tollbooth sensors provide accurate vehicle counts by category, while extensive mobility data from cellular network activity captures aggregated crowd movement, but lack modal disaggregation and have systematic biases.
    This study develops a machine learning framework to correct and disaggregate cellular network data using sparse tollbooth counts as ground truth. The model uses temporal and spatial features to learn the complex relationship between aggregated mobility data and vehicular data. The framework infers destinations from transit routes and implements routing logic to distribute corrected flows between OD pairs.
    This approach is applied to a bus depot expansion in Trondheim, Norway, generating hourly OD matrices by vehicle length category. The results show how limited but accurate sensor measurements can correct extensive but aggregated mobility data to produce grounded estimates of background vehicular traffic flows. These macro-scale estimates can be refined for micro-scale analysis at desired locations.
    The framework provides a generalisable approach for generating origin--destination data from cellular network data. This enables downstream tasks, like detailed traffic simulations for infrastructure planning in data-scarce contexts, supporting urban planners in making informed decisions.
\end{abstract}


\section{Introduction}
\label{sec:introduction}
Urban mobility is under pressure from population growth and urban expansion, making sustainable transport solutions essential to mitigate negative externalities such as traffic congestion, air pollution, and environmental impacts \cite{Ritchie2020cpt}. Furthermore, research indicates that sustainable modes (public transit and active mobility) result in relatively higher satisfaction and positive emotions among users while reducing the share of car-based trips \cite{Mouratidis2023stm}. With sustainable urban mobility representing a paradigmatic shift in urban planning goals towards creating liveable cities, public transit plays a central role in this transformation.

However, expanding public transit capacity requires understanding how increased coverage and service scheduling will interact with existing urban traffic flows. Traffic simulations enable planners to conduct ``What If?'' scenario analyses to identify bottlenecks and undesired deviations. This capability is crucial for planning infrastructure interventions, such as dedicated bus lanes, priority signals, or intersection redesigns. This study is motivated by the planned bus depot expansion in Trondheim, Norway, where planners must evaluate how increased bus operations will interact with background traffic flows. Without this analysis, unexpected interactions could lead to increased congestion around the depot, negatively impacting bus punctuality and service reliability.

The core challenge with creating traffic simulations is the lack of detailed origin--destination (OD) data by vehicle category. The bus scheduling data are readily available from the transit authority at the required granularity and are easily convertible to OD matrices. However, background traffic data presents the primary problem as there is no data source which tracks vehicles from origin to destination across the city. Two complementary but imperfect data sources are available. The Norwegian Public Roads Administration (NPRA) operates automated tollbooths on major roads that provide hourly vehicle counts disaggregated by length categories \cite{NPRA2023trafikkdata}. However, these tollbooths are sparsely located and provide point counts without OD information.

On the other hand, anonymised mobility data from Telia's ``Crowd Insights'' platform \cite{Telia2021cim} is available as routing reports that estimate crowd movements (\it{peopleFlow}) across the road network with extensive spatial coverage. However, these data lack disaggregation by mobility mode or vehicle category, suffer from systematic biases (e.g., overestimating traffic on major highways, inaccuracies from privacy-preserving aggregation), and provide flow volumes without OD information. Neither source alone provides the granular OD matrices per vehicle category required for traffic simulation software.

This paper proposes a methodology to fuse these heterogeneous data sources by leveraging their complementary strengths: sparse/accurate/disaggregated (tollbooth counts) versus extensive/noisy/aggregated (routing reports). The primary objective is to develop a machine learning (ML) framework that corrects and disaggregates cellular network data using sparse tollbooth counts as ground truth. The approach comprises three components: \it{(i)} defining the study area with OD locations based on tollbooth locations and inferred destinations, \it{(ii)} training a model to learn the relationship between aggregated mobility data and vehicular tollbooth counts while accounting for temporal and spatial biases, and \it{(iii)} implementing routing logic to distribute corrected and disaggregated flows between OD pairs. The methodology addresses the fundamental challenge of converting aggregated mobility data into actionable traffic simulation inputs.

The primary contributions of this study can be summarised as follows:

\begin{itemize}
    \item It presents a generalisable data fusion framework for estimating background traffic flows in data-scarce environments, enabling detailed traffic simulations for infrastructure planning.
    \item The framework demonstrates how sparse/accurate sensor data serve as ground-truth for extensive/noisy datasets, facilitating the correction and disaggregation of aggregated mobility data.
    \item It provides a practical case study of a planned bus depot expansion in Trondheim, Norway, generating vehicle-category-specific OD matrices required for traffic impact analysis.
    \item The approach enables macro-level traffic estimation that can be refined with targeted data collection at critical locations for richer micro-scale mobility analysis.
\end{itemize}

\section{Related Work}
\label{sec:related-work}
Recent research in OD matrix estimation has moved beyond fixed-location sensors towards fusing traditional counts with opportunistic digital traces. Dynamic OD models now integrate crowd-sourced data such as floating car data and venue activity patterns to capture the spatio-temporal flexibility of travel demand \cite{Castiglione2025ado}. Similarly, \cite{Fernandes2025eod} estimated network-level OD matrix for Helsinki's bus network by integrating sparse smartphone trajectories with Automated Passenger Count (APC) data. The OD matrices were validated against Telia ``Crowd Insights'' cellular data and demonstrated that accurate physical counts can anchor biased digital mobility data.

A key obstacle when using cellular data for traffic estimation is the presence of systematic biases in the underlying data collection processes. At the trip level, different data sources systematically exclude different populations and trip types, but fusing multiple biased datasets through a behaviourally informed model can mitigate these shortcomings \cite{Guan2025umb}. Our framework follows this principle by using tollbooth counts as ground truth to locally correct the cellular \it{peopleFlow} estimates, rather than applying a global scaling factor.

On the modelling side, ML methods for OD estimation and traffic correction include deep learning architectures such as knowledge-enhanced GNNs \cite{Xing2025odp} and GraphResLSTM models \cite{Hu2025odp}, as well as frameworks that embed optimisation within training loops for sensor error correction \cite{Zheng2025eem}. These approaches require dense, structured OD data that is unavailable in the data-scarce setting addressed here. In similar traffic prediction tasks, XGBoost has shown competitive accuracy with lower computational overhead than LSTM and Random Forest \cite{Lam2026tpu}. XGBoost is therefore adopted for the present study, prioritising interpretability and computational efficiency over deeper architectures.

The most closely related work is CTCam \cite{Lin2023cet}, which fuses cellular network data with sparse camera-based vehicle counts through a two-stage approach. However, CTCam focuses on enhancing cellular traffic prediction without disaggregating flows by vehicle category or producing OD matrices. The present study addresses this gap by extending the fusion paradigm to vehicle-category disaggregation and OD matrix generation.

\section{Methodology}
\label{sec:methodology}
This section details the process of combining routing reports with tollbooth data (as ground truth) to estimate background traffic flows around the Sandmoen bus depot in Trondheim, Norway. The methodology is broken down into three main components: defining the data sources and the study area, developing a routing logic for the flow distribution, and implementing a data fusion model to generate traffic estimates by vehicle category.

\begin{figure}[tb]
    \centering \includegraphics[width=\linewidth]{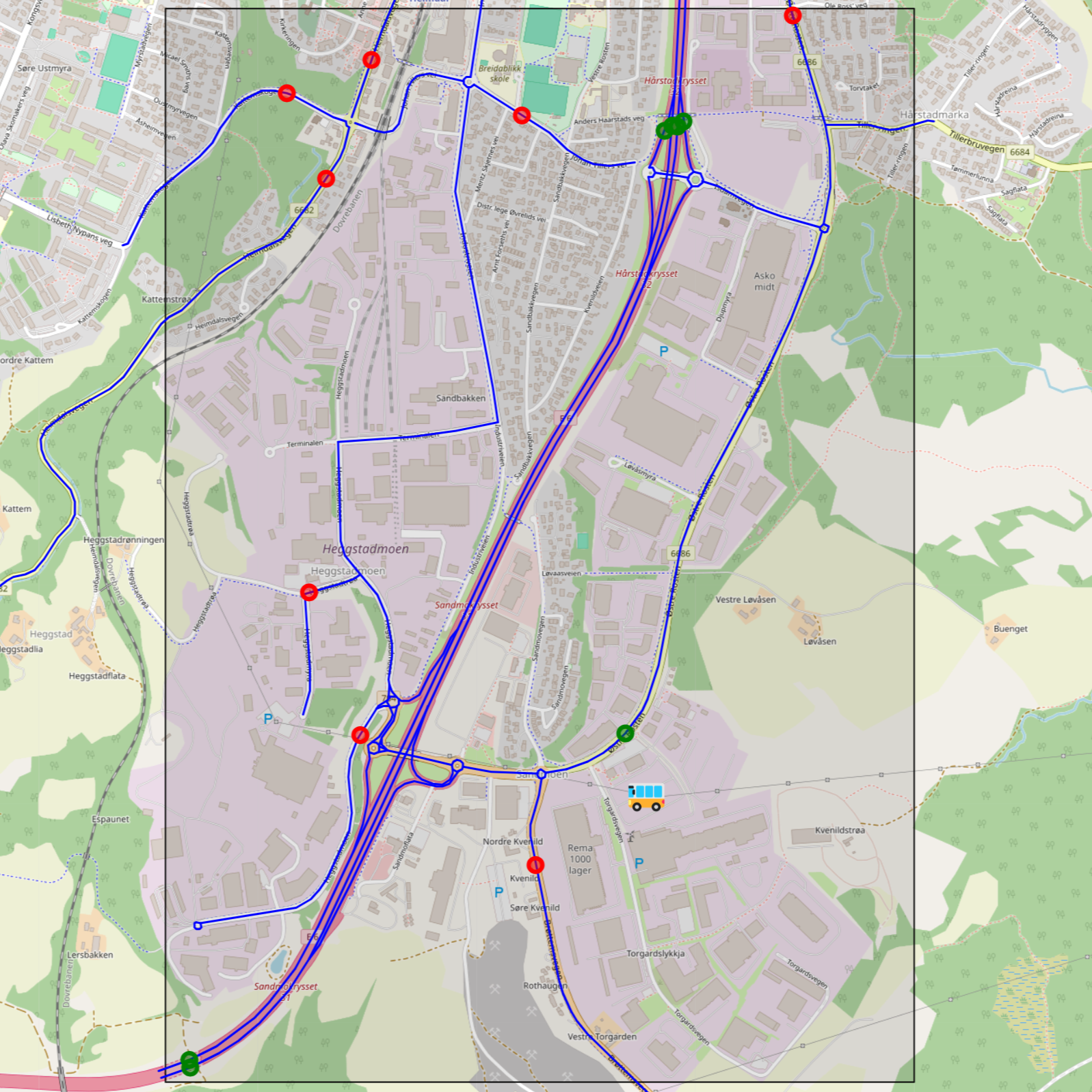}
    \caption{Map of the study area around the Sandmoen bus depot (bus icon), showing the coverage of routing reports (blue lines), NPRA tollbooths (green circles), and destinations inferred from bus routes (red circles).}
    \label{fig:study-area}
\end{figure}

\subsection{Data Sources and Study Area}
The study area, shown in Fig.~\ref{fig:study-area}, is centred around the bus depot (bus icon). Its boundaries are defined by the locations of the NPRA tollbooths (green circles) and the coverage of the mobility data derived from cellular network activity (blue lines). These sources provide complementary information: tollbooths offer accurate vehicle counts at fixed points, while cellular data provides extensive spatial coverage across the road network.

NPRA tollbooths provide accurate hourly vehicle counts disaggregated by length categories (e.g., under 5.6m, 5.6--7.6m, etc.). However, the tollbooths are sparsely located on major roads and lack origin--destination (OD) information. Five main tollbooths define the primary entry and exit points of the study area, supplemented by three county road tollbooths (outside the study area) selected to improve data overlap and coverage of less represented road types for model training.

The cellular network data provides routing reports with ``peopleFlow'' estimates of aggregated crowd movements across an extensive road network. These data offer broad spatial coverage, but lack disaggregation by mobility mode or vehicle type. The raw counts are post-processed by Telia to correct for telecom market share, ensuring the estimates reflect population-level mobility. Furthermore, the data processing assumptions lead to certain systematic biases: \it{(i)} traffic volumes can be misrepresented, leading to traffic on major highways being overrepresented (and vice-versa for smaller roads), and \it{(ii)} the minimum aggregation threshold adopted to preserve user privacy leads to underreported flows in low-traffic areas or times.

To create a complete network for traffic simulations, key destinations within the study area were inferred from bus scheduling data covering 64 bus lines. The analyses of probable routes from the depot to the first service stops initially identified seven routes and their corresponding destinations. The selection was refined based on the coverage of the routing reports, yielding eight destinations within the study area where the bus routes interact with the road network. The NPRA tollbooths and inferred destinations, summarised in Table~\ref{tab:tollbooths-destinations}, form the OD nodes for the estimation of the background traffic flows.

\begin{table}[tb]
    \renewcommand{\arraystretch}{1.15}
    \caption{Summary of Tollbooths and Inferred Destinations Used in the Study}
    \label{tab:tollbooths-destinations}
    \begin{center}
        \begin{tabular}{ll}
            \toprule
            \bt{Category}              & \bt{Name (Description)}            \\
            \midrule
            \bt{Main Tollbooths}       & E6--Klett (Passthrough)            \\
                                       & E6--Trondheim (Passthrough)        \\
                                       & Onramp--Isdamvegen (Inflow)        \\
                                       & Isdamvegen--Offramp (Outflow)      \\
                                       & ØstreRosten (County)               \\
                                       & Storlersbakken--Klett (Outflow)    \\
                                       & Storlersbakken--Trondheim (Inflow) \\
            \midrule
            \bt{Additional Tollbooths} & Bjørndalsbrua (County)             \\
                                       & Granåsen (County)                  \\
                                       & Heimdalsvegen (County)             \\
            \midrule
            \bt{Inferred Destinations} & Brøttemsvegen                      \\
                                       & Heggstadmoen                       \\
                                       & Heggstadtrøa                       \\
                                       & Heimsdalvegen (Nord \& Sør)        \\
                                       & Johan Tillers veg                  \\
                                       & Kattemskogen                       \\
                                       & Østre Rosten (Nord)                \\
            \bottomrule
        \end{tabular}
    \end{center}
\end{table}

Due to data availability constraints, two distinct datasets were used for modelling and simulation purposes. While this introduces a temporal mismatch, it is assumed that the fundamental traffic distribution patterns by hour and vehicle length category remained relatively stable between these periods, as the underlying road network and typical usage patterns did not change significantly. The two datasets are summarised as follows:

\begin{itemize}
    \item \it{Modelling Dataset:} This dataset covers November 2023 and contains 8,712 hourly samples from all eight tollbooths (with directional splits for certain locations), combined with routing reports from the same period. It is used to train the data fusion model to learn the relationship between aggregated cellular data and vehicular traffic counts.
    \item \it{Simulation Dataset:} This dataset covers 30--31 January 2025 and uses only the five main tollbooths within the study area. It is used in combination with the routing logic and trained model to generate the final OD matrices for traffic simulation.
\end{itemize}

To support the stability assumption, \it{peopleFlow} temporal patterns at tollbooth locations were compared between 2019 (pre-pandemic) and 2023 (post-pandemic). Pearson correlation exceeded 0.99 for both diurnal and weekly distributions, with symmetric Kullback--Leibler divergence below 0.002 nats and normalised mean square error below 0.004, confirming structural stability across years. This justifies using the 2023 modelling dataset to predict 2025 traffic patterns.

\begin{figure*}[tb]
    \centering \includegraphics[width=0.90\linewidth]{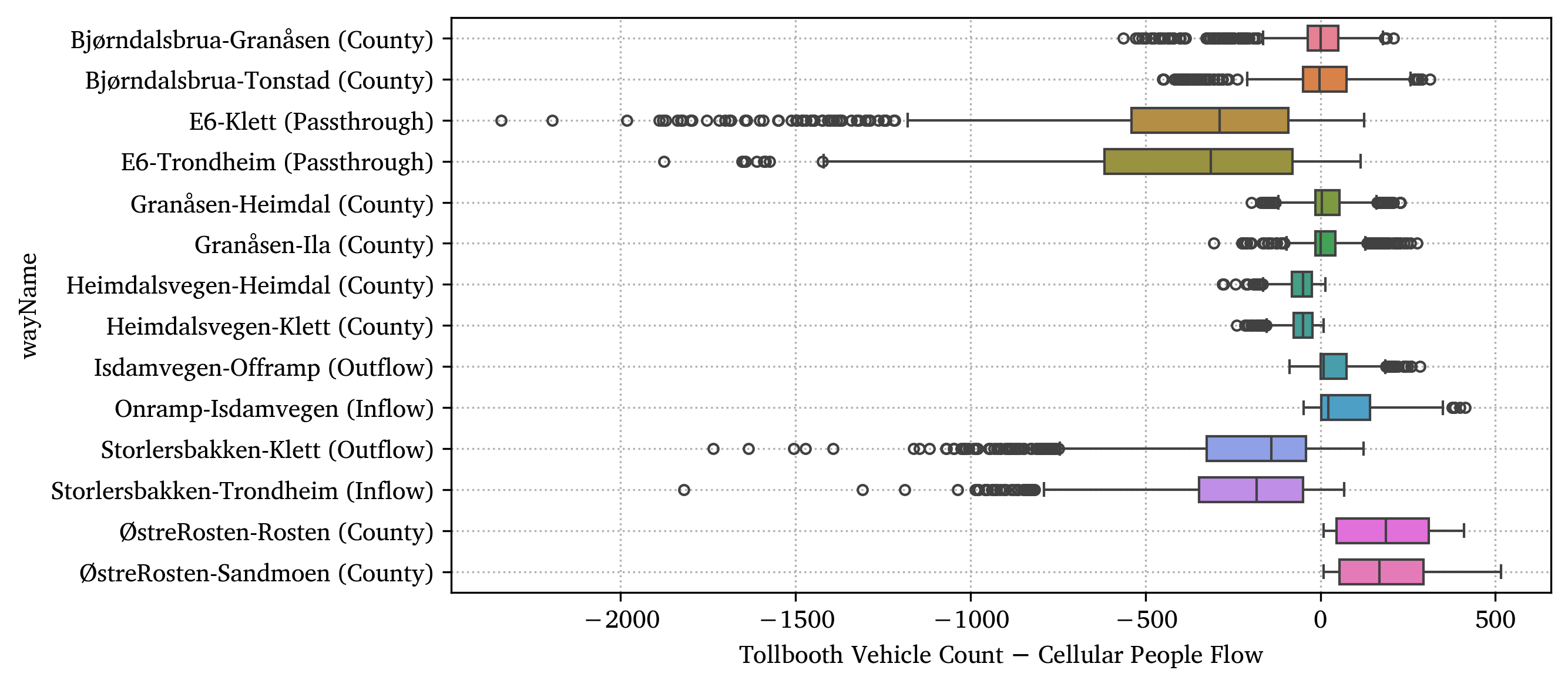}
    \caption{Differences between total vehicular counts and \it{peopleFlow} at tollbooth locations, aggregated by mean across the ``modelling'' dataset. \it{Note: Positive values indicate that tollbooth counts are higher than cellular data estimates.}}
    \label{fig:data-comparison}
\end{figure*}

\subsection{Routing Logic}
With the network of tollbooths (boundaries) and destinations (internal nodes) established, a routing logic was manually developed to distribute traffic flows within the study area. This logic reconstructs plausible traffic routes by interpreting hourly imbalances between tollbooth counts upstream and downstream of the bus depot as net inflows or outflows. The logic handles three primary traffic scenarios sequentially:

\begin{enumerate}[itemsep=0.5em,topsep=0.5em]
    \item \it{Internal Flows:} Traffic circulating within the study area is handled first. The Østre Rosten tollbooth acts as a boundary between two local sub-regions. By comparing flows at this tollbooth with traffic on the nearby Isdamvegen ramps, the logic determines the net direction of local traffic (e.g., eastbound or westbound) and assigns the surplus flow between internal destination groups.
    \item \it{Local Inflows/Outflows:} After accounting for internal circulation, any remaining traffic on the Isdamvegen ramps is treated as entering (inflow) or exiting (outflow) the study area. These flows connect the main highway with the study area and are routed between the ramps and the inferred internal destinations.
    \item \it{Passthrough Flows:} Finally, traffic on the main E6 highway is processed. By comparing counts at paired tollbooths north and south of the study area (e.g., ``E6--Klett'' and ``Storlersbakken--Klett''), the logic determines the net inflow or outflow for that hour and routes the difference to or from the inferred destinations. The minimum shared volume between the two tollbooths is treated as passthrough traffic that bypasses the local network entirely.
\end{enumerate}

This structured logic provides the basis on which the corrected and disaggregated vehicle counts from the data fusion model are assigned to specific OD pairs. However, it simplifies reality by assigning traffic in one dominant direction per hour (e.g., northbound or southbound), a limitation acknowledged in this study.

\subsection{Data Fusion}
Ideally, vehicular traffic should be a subset of urban mobility, meaning the total tollbooth count should be less than the \it{peopleFlow} estimate. However, as shown in Fig.~\ref{fig:data-comparison}, the relationship is more complex as the differences between the two data sources vary significantly by time of day and type of road. The systematic biases in the cellular network data previously discussed are inconsistent, making simple statistical operations infeasible for correcting/disaggregating the cellular data using the tollbooth counts. The differences fluctuate between positive and negative values across the ``modelling'' dataset, with positive differences indicating that cellular data sometimes underestimate traffic, and vice-versa.

\begin{figure*}[tb]
    \centering \includegraphics[width=0.75\linewidth]{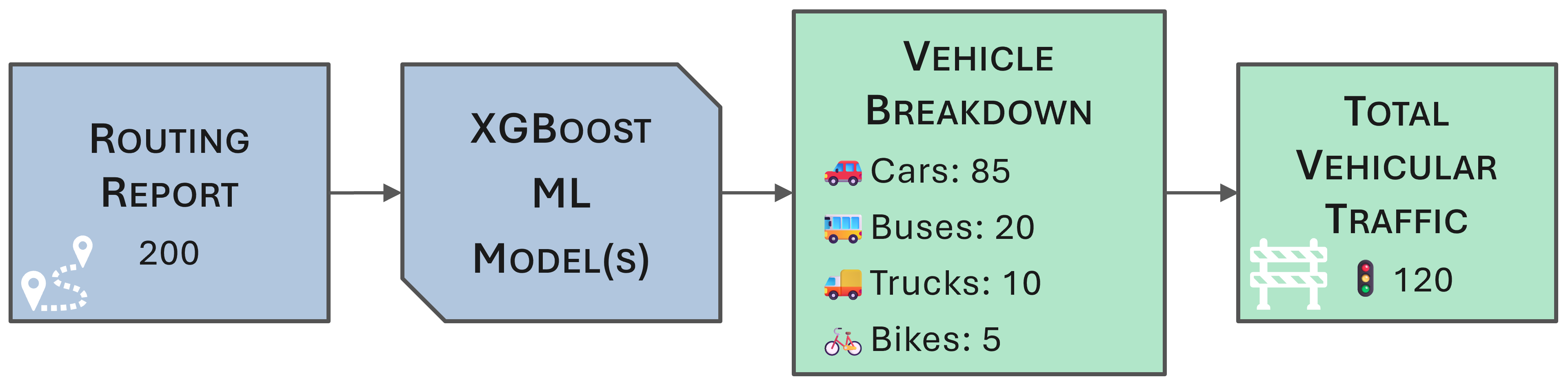}
    \caption{The ML-based pipeline adopted for data fusion. The XGBoost model is trained to learn the relationship between aggregated \it{peopleFlow} and vehicular traffic counts at tollbooth locations, using temporal and spatial features to account for systematic biases.}
    \label{fig:xgb-pipeline}
\end{figure*}

To address this complexity, the ML-based pipeline illustrated in Fig.~\ref{fig:xgb-pipeline} was developed to correct and disaggregate cellular data. Given aggregated \it{peopleFlow} volumes for tollbooth locations, a XGBoost model \cite{Chen2016xas} was trained on the ``modelling'' dataset to predict the total and length-specific vehicle counts with the tollbooth data as ground truth. The model uses the following input features:

\begin{itemize}
    \item \it{peopleFlow:} The raw mobility volume from the routing reports, representing aggregated crowd movement derived from cellular data.
    \item \it{Temporal Features:} Hour of the day (0--23), day of the week, and a weekend/weekday flag to capture daily and weekly traffic patterns.
    \item \it{Spatial Features:} A categorical tag for the road type (Primary, Trunk, Secondary) to account for systematic biases varying by road hierarchy.
\end{itemize}

The trained model learns the complex relationship between aggregated \it{peopleFlow} and vehicular traffic at the tollbooth locations. It is subsequently applied to the routing reports for the study area from the ``simulation'' dataset, including the inferred destinations where no ground-truth data exists. This process produces hourly flow estimates per vehicle length category for each destination, which are subsequently converted into probability distributions for use with the routing logic.

\section{Results and Discussion}
\label{sec:results-discussion}
This section evaluates the performance of the data fusion model, outlines the process of generating the origin--destination (OD) matrices by applying the trained model and routing logic to the ``simulation'' dataset, and discusses the integration of the generated OD matrices into traffic simulations.

\subsection{Model Evaluation}
Table~\ref{tab:model-performance} summarises the performance of the XGBoost model on the validation subset of the ``modelling'' dataset with the raw \it{peopleFlow} as baseline. The uncorrected \it{peopleFlow} achieves a validation $R^2$ of only 0.4437, confirming its inadequacy as a direct proxy for vehicular traffic if used as-is. In contrast, the XGBoost model achieves $R^2$ of 0.9812 for total traffic volume and $R^2 > 0.97$ per-vehicle category. These results demonstrate the model's effectiveness in correcting and disaggregating the routing reports to produce accurate vehicle count estimates.

\begin{table}[ht!]
    \renewcommand{\arraystretch}{1.15}
    \centering
    \caption{XGBoost Model Performance Summary, with Uncorrected \textnormal{\it{peopleFlow}} as Baseline}
    \label{tab:model-performance}
    \begin{tabular}{lcc}
        \toprule
        \bt{Vehicle Length}  & \bt{Train (RMSE / $R^2$)} & \bt{Valid (RMSE / $R^2$)} \\
        \midrule
        \it{peopleFlow}      & 279.67 / 0.5072           & 293.09 / 0.4437           \\
        Total Traffic Volume & \bt{59.98 / 0.9841}       & \bt{64.63 / 0.9812}       \\
        \midrule
        $<5.6$ m             & 59.24 / 0.9799            & 63.60 / 0.9763            \\
        5.6--7.6 m           & 3.84 / 0.9984             & 4.28 / 0.9979             \\
        7.6--12.5 m          & 5.85 / 0.9958             & 6.90 / 0.9941             \\
        12.5--16.0 m         & 2.90 / 0.9991             & 3.19 / 0.9990             \\
        16.0--24.0 m         & 3.63 / 0.9986             & 3.94 / 0.9983             \\
        $\geq 24.0$ m        & 1.50 / 0.9998             & 1.57 / 0.9998             \\
        \bottomrule
    \end{tabular}
\end{table}

\begin{figure*}[tb]
    \centering
    \begin{minipage}[t]{0.49\linewidth}
        \includegraphics[width=\linewidth]{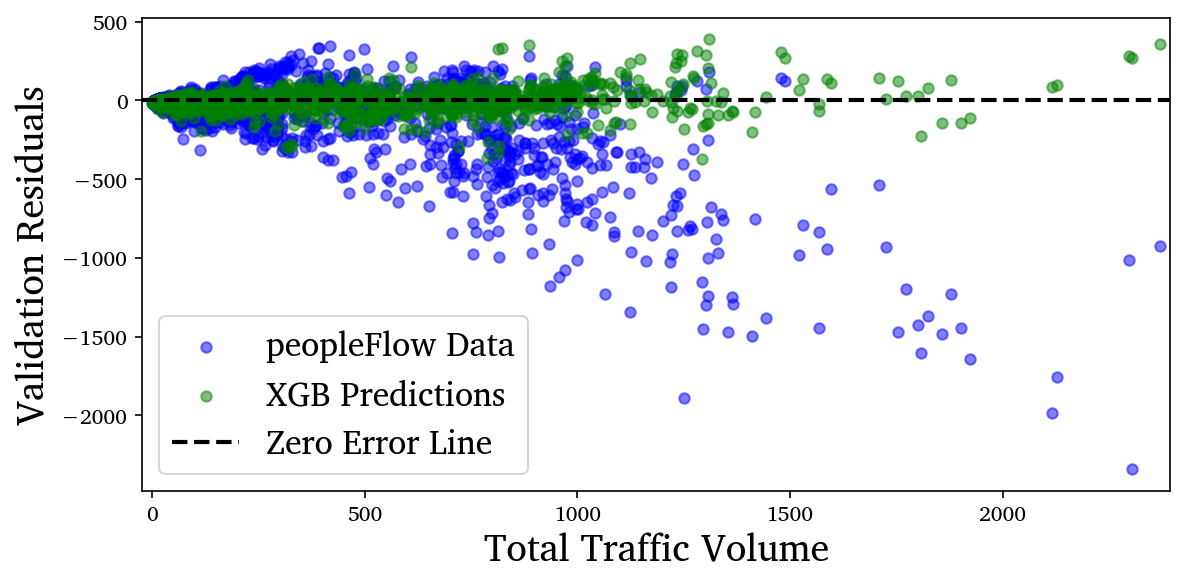}
        \caption{Comparison of prediction residuals for total traffic volume between the raw \it{peopleFlow} data and the trained XGBoost model. The raw data's residuals increase with traffic volume, while the model effectively corrects these biases.}
        \label{fig:xgb-residuals}
    \end{minipage}
    \hfill
    \begin{minipage}[t]{0.49\linewidth}
        \includegraphics[width=\linewidth]{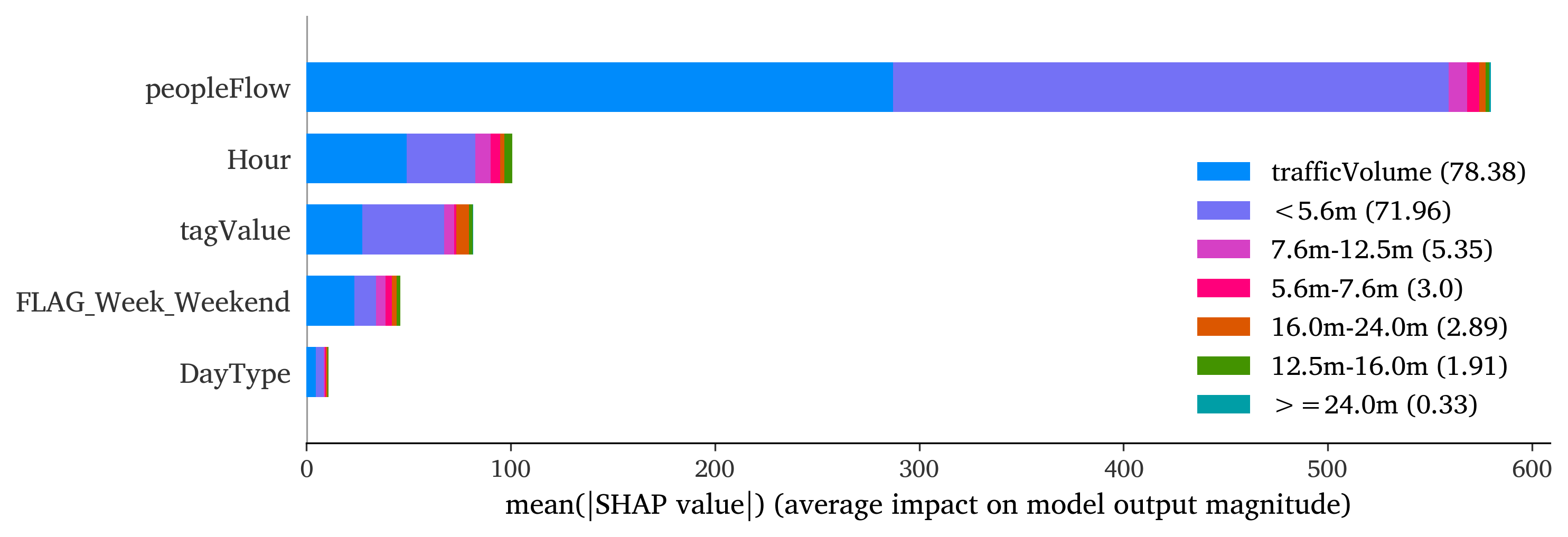}
        \caption{SHAP summary plot showing the global importance of each feature to the XGBoost model's predictions. \it{peopleFlow}, \it{Hour}, and \it{tagValue} are the most influential features overall.}
        \label{fig:xgb-shap}
    \end{minipage}
\end{figure*}

In addition to aggregate metrics, the model's ability to correct systematic biases in the cellular data was similarly examined through residual analysis. Fig.~\ref{fig:xgb-residuals} compares the prediction residuals of the raw \it{peopleFlow} data with those of the XGBoost model predictions for the total traffic volume. While uncorrected \it{peopleFlow} residuals increase significantly with traffic volume, the model's residuals remain stable and close to the zero-error line throughout the range. This demonstrates that the model successfully mitigates the systematic under- and over-estimation biases present in the cellular data.

\begin{figure*}[htbp!]
    \centering \includegraphics[width=0.91\linewidth]{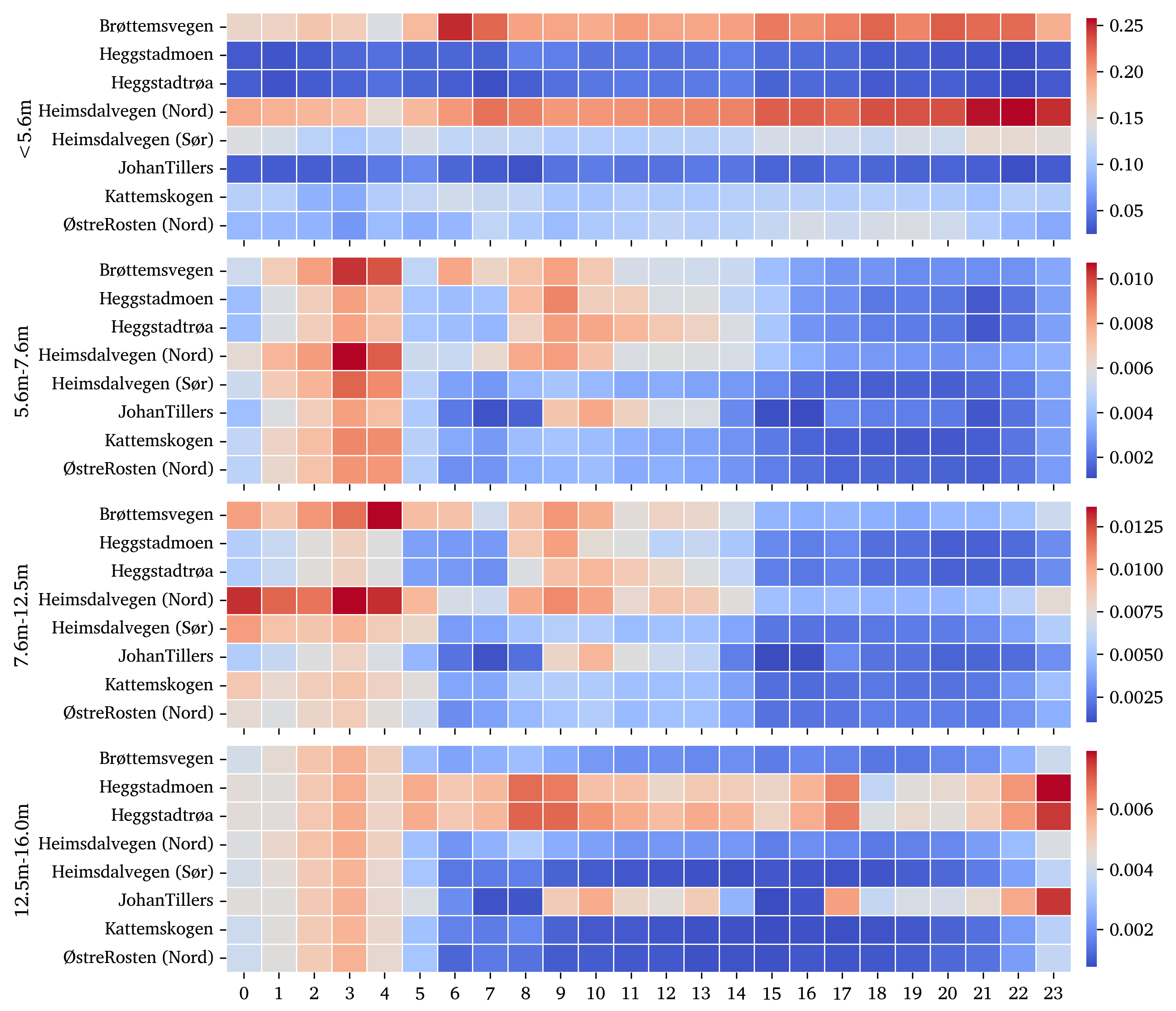}
    \caption{Hourly inferred traffic flow distributions across the eight destinations for various NPRA vehicle length categories, generated by applying the XGBoost model to routing reports from November 2023. \it{Note: For each hour, the distribution sums to 1 across all destinations and vehicle categories combined, representing a joint probability over (destination, vehicle category) pairs.}}
    \label{fig:inferred-distributions}
\end{figure*}

\begin{figure*}[tb]
    \centering
    \begin{subfigure}{0.32\linewidth}
        \includegraphics[width=\linewidth]{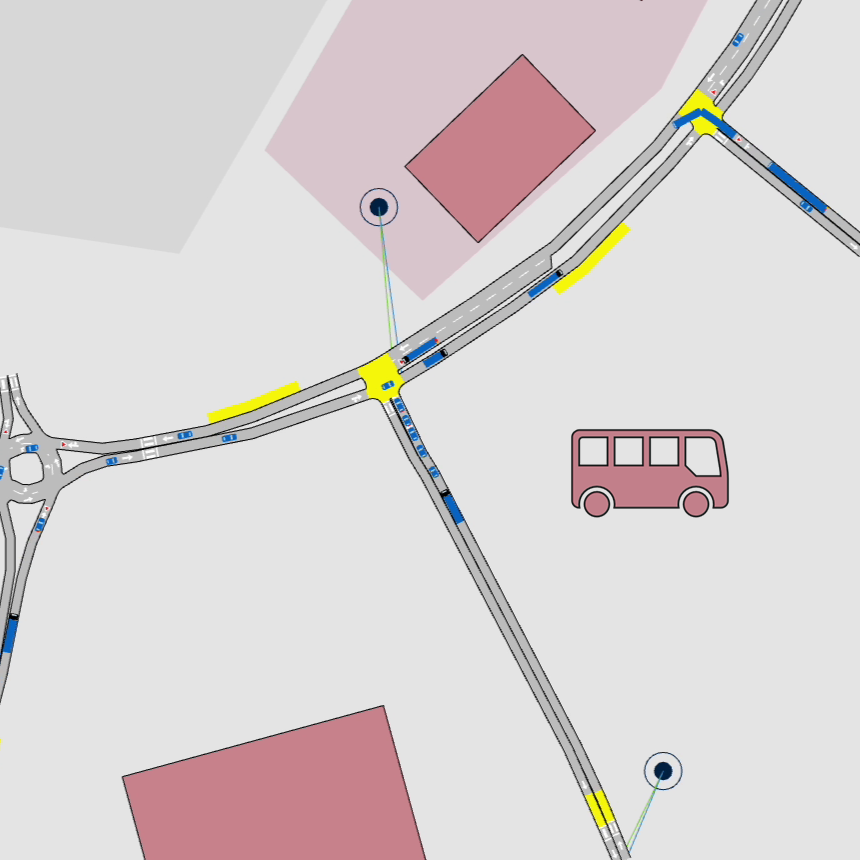}
        \caption{Reference Scenario}
    \end{subfigure}
    \begin{subfigure}{0.32\linewidth}
        \includegraphics[width=\linewidth]{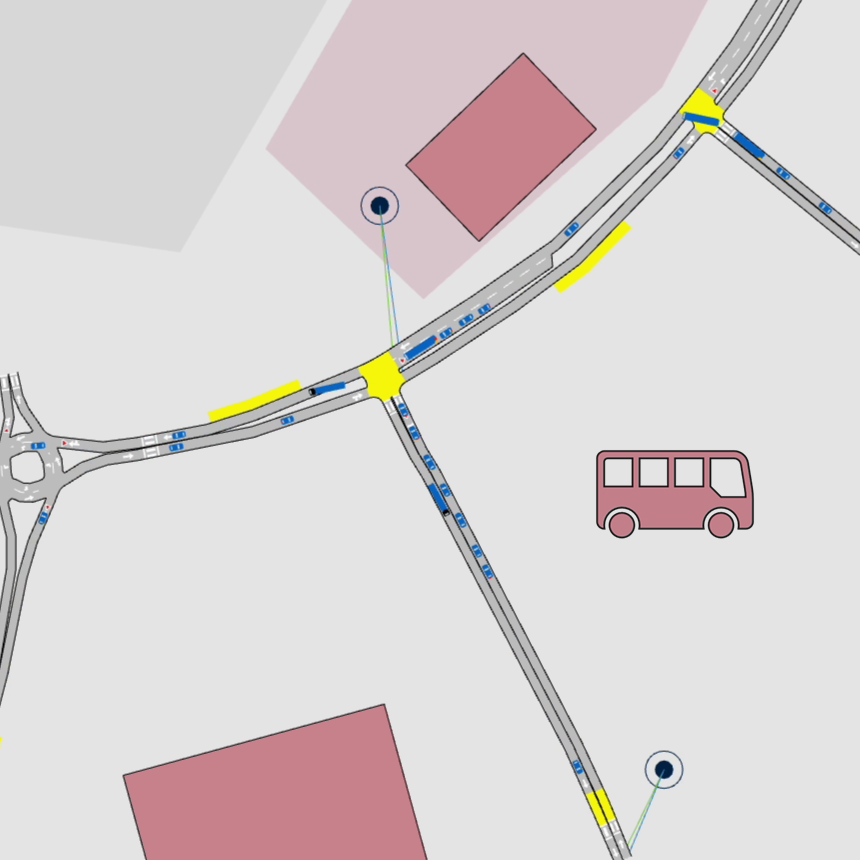}
        \caption{Medium Growth Scenario}
    \end{subfigure}
    \begin{subfigure}{0.32\linewidth}
        \includegraphics[width=\linewidth]{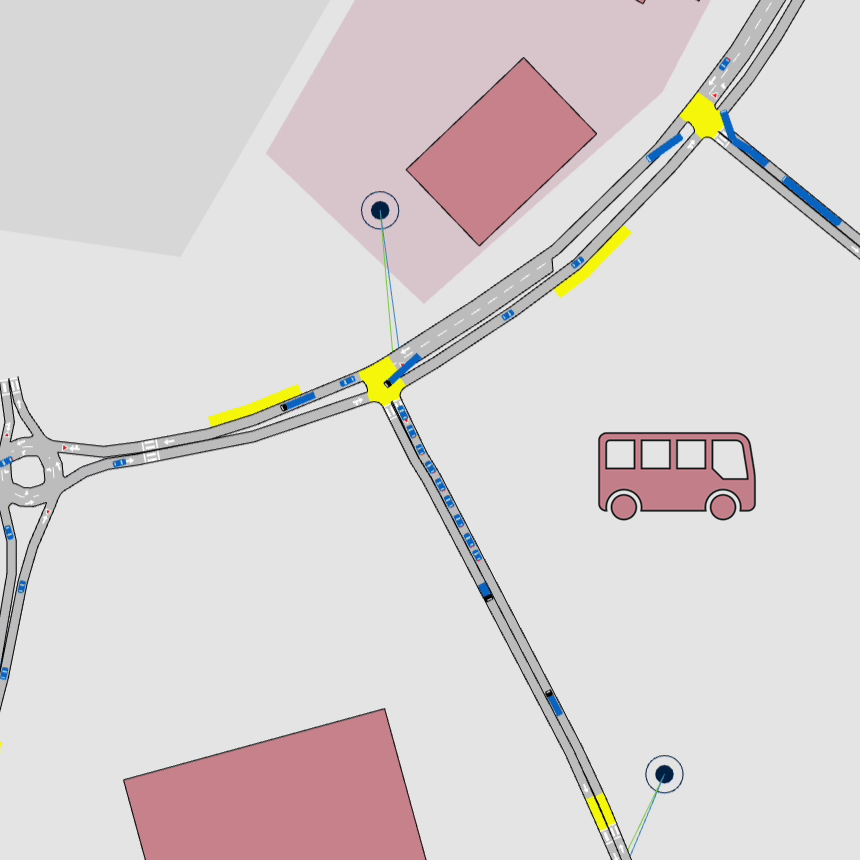}
        \caption{High Growth Scenario}
    \end{subfigure}
    \caption{Outputs from the Aimsun traffic simulation, showing (a) the reference scenario with estimated background traffic and planned bus schedules, (b) a medium growth scenario with increased bus frequency, and (c) a high growth scenario with further increased bus frequency.}
    \label{fig:aimsun-simulation-scenarios}
\end{figure*}

Furthermore, we investigated the importance of input features to the model's predictions using SHapley Additive exPlanations (SHAP) \cite{Lundberg2017aua}, as shown in Fig.~\ref{fig:xgb-shap}. The \it{peopleFlow} feature is the most influential predictor, as expected. The temporal \it{Hour} and spatial \it{tagValue} (type of road) features are also quite significant, confirming that the traffic patterns vary by time of day and road category. In particular, the weekend/weekday flag proved more influential than individual days of the week, indicating that weekday-weekend differences dominate over day-specific patterns.

Taken together, the feature importance analysis underscores the initial observation (see Fig.~\ref{fig:data-comparison}) that the systematic biases present in the cellular network data exhibit considerable spatiotemporal variation. The inclusion of additional temporal and spatial features proved effective by allowing the model to make contextual adjustments to the \it{peopleFlow} input for better predictive performance, as demonstrated by the $R^2$ scores and residual analysis.

\subsection{Origin--Destination Matrices}
The trained XGBoost model was applied to generate hourly vehicular traffic flow estimates for the inferred destinations where no tollbooths counts were available. These estimates were converted into probability distributions for allocating traffic flows between tollbooths and destinations for each vehicle length category based on the routing logic. For each hour, the complementary roles of the routing logic and data fusion model can be summarised in the following steps:

\begin{enumerate}[itemsep=0.5em,topsep=0.5em]
    \item \it{Model Inference on Cellular Data:} The XGBoost model is applied to the routing reports at the eight inferred destinations, producing corrected flow estimates for each vehicle length category. These predictions are normalised into a joint probability distribution over (destination, vehicle category) pairs for each hour. Marginalising over vehicle categories yields a \it{global distribution} representing relative traffic share across destinations, while marginalising over destinations yields \it{destination-specific distributions} for vehicle category breakdown.
    \item \it{Determine Total Flows and Direction:} Using the ``simulation'' dataset, the routing logic compares tollbooth counts upstream and downstream of the bus depot to determine the total volume and the dominant direction of traffic. For example, if the upstream tollbooth records 500 vehicles and the downstream records 400, then 100 vehicles are classified as inflow to the study area, while the remaining 400 are passthrough traffic.
    \item \it{Distribute Flows Across Destinations:} The routing logic determines which subset of destinations is affected by each traffic scenario (e.g., internal flows only involve destinations within certain sub-regions). The global distribution is renormalised for this subset and is used to allocate the flow. For instance, if 50 of the 100 inflow vehicles are routed to a subset containing Brøttemsvegen and Heimsdalvegen, and these account for 60\% and 40\% of the renormalised distribution, the vehicles are split as 30 and 20, respectively.
    \item \it{Disaggregate by Vehicle Category:} Finally, each destination's specific distribution is applied to determine the vehicle category breakdown. For example, the 30 vehicles routed to Brøttemsvegen might consist of 22 passenger cars, 5 light commercial vehicles, and 3 trucks.
\end{enumerate}

Since converting continuous probability distributions to discrete vehicle counts involves rounding, a largest remainder method is used during flow distribution to ensure that the total is neither over- nor under-allocated and no destinations are left without assigned vehicles. The aforementioned steps are applied for each hour and traffic scenario (internal, local inflow/outflow, passthrough), populating the OD matrix with entries specifying origin, destination, vehicle type, and count.

Fig.~\ref{fig:inferred-distributions} visualises the joint probability distributions generated by the XGBoost model for the higher-volume vehicle categories. Each subplot shows the hourly distribution across the eight inferred destinations for a specific vehicle length category, with the colour intensity representing the probability mass allocated to each destination. The distributions reveal consistent patterns in traffic allocation, with certain destinations such as \it{Brøttemsvegen} and \it{Heimsdalvegen Nord} consistently receiving higher shares of traffic. This pattern reflects the underlying road hierarchy and connectivity within the study area.

\subsection{Traffic Flow Simulation}
While detailed simulation analysis is beyond the scope of this paper, the downstream task was used to validate the methodology presented here. To facilitate the integration of the generated OD matrices with simulation software, the NPRA vehicle length categories were mapped to standard vehicle types used in the relevant literature \cite{Gaynor2023aev}, as follows:

\begin{itemize}
    \item \it{Passenger Vehicles} [under 5.6m]: Includes private cars, taxis, and SUVs.
    \item \it{Light Commercial Vehicles} [5.6--7.6m]: Includes larger passenger vans, pickup trucks, and delivery vans.
    \item \it{Buses \& Medium Trucks} [7.6--12.5m]: Includes single-unit medium-duty trucks, city buses, and coaches.
    \item \it{Heavy Rigid \& Short Articulated Trucks} [12.5--16.0m]: Includes heavy rigid trucks and shorter articulated combinations.
    \item \it{Articulated Heavy Goods Vehicles} [16.0--24.0m]: Includes full-length tractor-trailer combinations and other long heavy goods vehicles.
    \item \it{Extra-Long Vehicles} [over 24.0m]: Special category for very long combinations such as multi-trailer road trains and other oversized rigs.
\end{itemize}

The estimated background traffic flows were combined with OD matrices derived from the bus scheduling data for January 2025 and imported into the Aimsun traffic simulation software \cite{Aimsun2026ant}. Fig.~\ref{fig:aimsun-simulation-scenarios} depicts three scenarios explored: \it{(i)} a reference scenario with current bus operations (160 buses, 322 trips/day) and normal background traffic levels (kept constant); \it{(ii)} a medium growth scenario with increased bus frequency (180 buses, 360 trips/day); and \it{(iii)} a high growth scenario with further service expansion (360 buses, 720 trips/day).

The initial simulation results \cite{Tabassum2025hmb} revealed that the estimated OD flows for the two critical junctions adjacent to the depot (within the orange box in Fig.~\ref{fig:study-area}) were still underestimated. To improve accuracy, supplementary OD data were collected during rush-hour periods by deploying cameras at these junctions and performing hourly vehicle counts. The resulting junction-level OD information was integrated into the existing simulation framework to refine the flow estimates. This highlights a key strength of our proposed approach: the ability to generate macro-scale traffic flow estimates that can be refined with targeted data collection for micro-scale analysis at critical locations.

\section{Conclusion and Future Work}
\label{sec:conclusion}

\subsection{Conclusion}
This paper presented a ML-based data fusion framework to address the challenge of generating origin--destination (OD) data for traffic simulation in data-scarce environments. By fusing sparse, high-accuracy tollbooth counts with extensive but noisy mobility data derived from cellular network activity, our methodology successfully corrects and disaggregates aggregated crowd movements into granular, hourly traffic flow estimates by vehicle category.

The framework was applied to the practical case of a planned bus depot expansion in Trondheim, Norway, producing the OD matrices required for traffic impact analysis. The results demonstrate that our approach provides a robust and generalisable solution for creating grounded traffic models where traditional data collection is infeasible. A key contribution is the concept of generating a macro-scale estimation of traffic flows that can be strategically refined with targeted, micro-level data collection at critical locations, offering a scalable and cost-effective workflow for urban planners.

\subsection{Limitations and Future Work}
Due to data availability constraints, the modelling dataset (November 2023) and simulation dataset (January 2025) are not temporally aligned. However, analysis of \it{peopleFlow} temporal patterns demonstrates that fundamental traffic distribution patterns remained relatively stable across years. The validity of the approach will be further strengthened by ensuring full temporal alignment between datasets in future work. In addition, the current routing logic was manually constructed and simplifies traffic by assigning a single dominant direction per hour. This could be enhanced by adopting a probabilistic routing approach or an automated network flow optimisation that captures bidirectional flows more realistically.

The framework itself is highly extensible. The data fusion modelling accuracy could be improved by incorporating advanced deep learning algorithms or additional data sources. Furthermore, uncertainty quantification could be integrated to provide confidence intervals for the estimated flows, which would improve the interpretability and robustness of the results. \hide{UQ can be implemented through prediction intervals or ensemble-based bounds to propagate confidence estimates from the ML model through the routing logic to the final OD matrices.} The methodology can be generalised to other regions or adapted for other mobility modes, such as pedestrian traffic, by analysing the residual \it{peopleFlow} after accounting for vehicular movements. This offers a pathway to developing holistic, multi-modal urban mobility models.

\section*{Acknowledgment}
This research received funding from the PERSEUS Doctoral Program, supported under the Marie Skłodowska-Curie grant agreement No. 101034240. The authors also acknowledge MobilitetsLab Stor-Trondheim for their financial contribution. The authors thank AtB AS (Trondheim's public transit authority) for supporting the study and providing the bus scheduling and supplementary OD data, and the Norwegian Public Roads Administration for access to the tollbooth data, which proved essential for the analysis.



\end{document}